\pdfoutput=1

\documentclass[11pt]{article}

\usepackage[]{EMNLP2022}

\usepackage{times}
\usepackage{latexsym}

\usepackage[T1]{fontenc}

\usepackage[utf8]{inputenc}

\usepackage{microtype}

\usepackage{inconsolata}
\usepackage{amsmath}
\usepackage{amssymb}
\usepackage{graphicx}
\usepackage{subcaption}
\usepackage{booktabs}
\usepackage{float}

\newcommand{\ourmodule}{MANTa}
\newcommand{\ourmodel}{MANTa-LM}
\newcommand{\rs}{$\!${\color{red}|}$\!$} 
\newcommand{\nors}{$\!${\color{white}|}$\!$} 

\title{\protect\ourmodule{}: Efficient Gradient-Based Tokenization\\ for Robust End-to-End Language Modeling}

\author{
    Nathan Godey\thanks{~~Equal contribution.}$\mkern6mu^{1,2}$ \quad Roman Castagné$^{*1,2}$ \quad Éric de la Clergerie$^1$ \quad Benoît Sagot$^1$ \\
    $^1$Inria, Paris, France \\
    $^2$Sorbonne Université, Paris, France \\
    \texttt{\{nathan.godey,roman.castagne,eric.de\_la\_clergerie,benoit.sagot\}@inria.fr}
}

\begin{document}
\maketitle

\begin{abstract}
Static subword tokenization algorithms have been an essential component of recent works on language modeling. However, their static nature results in important flaws that degrade the models' downstream performance and robustness. In this work, we propose \ourmodule{}, a \textbf{M}odule for \textbf{A}daptive \textbf{N}eural \textbf{T}okeniz\textbf{A}tion. \ourmodule{} is a differentiable tokenizer trained end-to-end with the language model. The resulting system offers a trade-off between the expressiveness of byte-level models and the speed of models trained using subword tokenization. In addition, our tokenizer is highly explainable since it produces an explicit segmentation of sequences into blocks. We evaluate our pre-trained model on several English datasets from different domains as well as on synthetic noise. We find that MANTa improves robustness to character perturbations and out-of-domain data. We then show that MANTa performs comparably to other models on the general-domain GLUE benchmark. Finally, we show that it is considerably faster than strictly byte-level models.
\end{abstract}

\section{Introduction}
In order to improve Language Models (LMs), the Natural Language Processing field has removed most of the system-induced biases in the last few years. For instance, practices that were once standard such as lemmatization, stemming and feature engineering have progressively disappeared in favor of general architectures trained on huge amounts of data, learning end-to-end which features may be leveraged to attain better performances. However, one essential part of LMs has seen little evolution: tokenization. Tokenizers convert sequences of characters into sequences of tokens (substrings of smaller length) which can then be embedded by the model. Subword tokenization algorithms~\cite{sennrich-etal-2016-neural,wu2016google,kudo-2018-subword} are a specific class of tokenizers designed in such a way that almost every string can be encoded and decoded with very few out-of-vocabulary tokens. They are used in the vast majority of recent LMs, but have been an essential part of NLP systems since much longer~\cite{mielke2021between}.

\begin{figure}
\centering
\includegraphics[width=\linewidth]{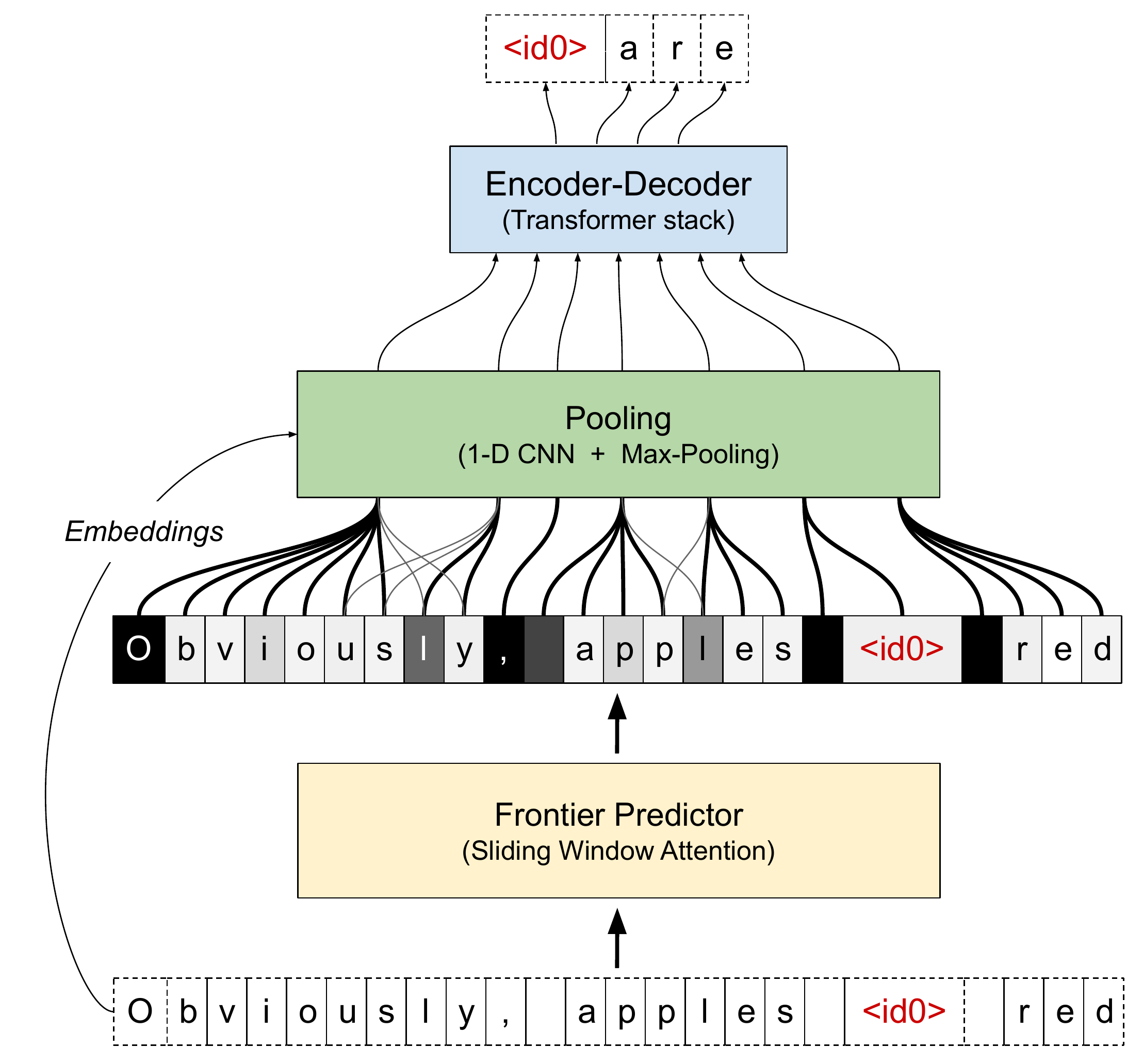}
\caption{The differentiable tokenization scheme of \ourmodel{}. Input bytes are first assigned a \textit{separation probability} using a Sliding Window Attention Transformer. These probabilities are used to compute the contribution of each byte embedding in the pooled representations of the \textit{blocks}. The block embeddings are fed to the Encoder-Decoder layers which predict the masked bytes. All the components are optimized with the LM objective.}
\label{fig:overview_diagram}
\end{figure}

The success of these algorithms can be attributed to several reasons. Firstly, they produce token sequences whose length is greatly reduced compared to the original character sequence. This characteristic is helpful because limitations in compute power and architectural constraints, such as the quadratic complexity with respect to sequence length of Transformers~\cite{vaswani2017attention}, prevent models from processing arbitrary long sequences. Secondly, they compress the corpus using occurrence statistics that may help LMs. For instance, if a word appears frequently in the training corpus, it will be encoded as a single token in the vocabulary and the model will be able to build a representation for that particular token more easily.

However, the induced biases of tokenizers are also harmful for modelization. One such limitation lies in their brittleness to character deformations which are commonly found in real world, noisy data. For instance, BERT's tokenizer~\cite{devlin-etal-2019-bert} encodes ``performance'' as \texttt{[``performance'']} but \mbox{``perfonmance''} as \texttt{[`per', `\#\#fo', `\#\#n', `\#\#man', `\#\#ce']}, which makes it hard for the model to behave similarly in both cases. Moreover, the tokenizer is fixed after its training and is therefore impossible to update, for instance to reflect new domains~\cite{el-boukkouri-etal-2020-characterbert} where tokenization might over-segment specific or technical terms. \citet{clark2022canine} list other issues emerging when using static subword tokenizers, especially when modeling languages with a more complex morphology than English.

To overcome these issues, \textit{tokenization-free} models~\cite{clark2022canine,xue2022byt5,tay2021charformer} produce character-based or byte-based embeddings for LMs instead of subword embeddings. These methods improve the robustness of LMs to naturally occurring noise as well as their expressiveness when dealing with out-of-domain or multilingual data. In order to cope with increased input lengths, some of these methods compress sequences with constant reduction rates obtained using specialized modules~\cite{clark2022canine,tay2021charformer}, subsequently removing any notion of subwords.

We argue that learning a subword tokenization together with input representations in an end-to-end fashion is beneficial for language modeling. In this work, we introduce \ourmodule{}, a gradient-based tokenizer and embedding module. It can easily be plugged-in to replace the classical combination of fixed tokenizers and trainable subword embedding matrices existing in most encoder-decoder models, without any increase in the total number of trainable parameters. We also introduce \ourmodel{}, a Transformer encoder-decoder that incorporates \ourmodule{} and that is trained end-to-end. By learning a soft, adaptive segmentation of input sequences jointly with the LM pre-training objective, \ourmodel{} produces byte-based representations with sequence lengths similar to those produced by static subword tokenizers. Additionally, by propagating gradients through our soft segmentation module during fine-tuning as well, we are able to adapt the segmentation to new domains, removing the limitations imposed by static subword tokenization.

We show that \ourmodel{} is robust to noisy text data and able to adapt to new domains while being significantly faster than byte-level models. Interestingly, \ourmodule{} learns a simple but explainable segmentation using only the LM objective while effectively reducing the length of byte sequences.

In summary, the contributions of this paper are the following: 
\begin{itemize}
    \item We introduce \ourmodule{}, a gradient-based tokenization and pooling module that can learn jointly with an encoder-decoder LM;
    \item We train \ourmodel{} on English data and we evaluate its robustness to synthetic and natural variation and its ability to adapt to new domains compared to byte-level models.
\end{itemize}

\section{Related Work}
Non-neural subword-level tokenization methods have dominated in the last few years as the default way to encode textual data, the most used being BPE \citep{sennrich-etal-2016-neural}, WordPiece~\cite{wu2016google} and Unigram \citep{kudo-2018-subword}. However, they have inherent flaws that limit their multilingual performance~\citep{rust-etal-2021-good}, their adaptability to new languages and new domains after pre-training~\citep{el-boukkouri-etal-2020-characterbert,garcia-etal-2021-towards} and the downstream performance of language models in general \citep{bostrom-durrett-2020-byte}.

To alleviate these issues, tokenization-free (or character-level) models leverage characters instead of subwords to build text representations. Some of the first neural networks for sequence generation used characters directly as inputs \citep{sutskever2011generating,graves2013generating}, and following works modified the approach to create input word representations based on characters \citep{kim2016character,Jzefowicz2016ExploringTL,peters-etal-2018-deep}. Similar architectures were recently adapted to work with Transformers \citep{el-boukkouri-etal-2020-characterbert,ma-etal-2020-charbert}. Nevertheless, they still rely on fixed tokenization heuristics (for instance segmenting using whitespaces) which may not be suited to some languages or certain types of language variations. Recent works have tried to remove these induced biases by working purely with characters or bytes as input \citep{clark2022canine,tay2021charformer,xue2022byt5}. However, they either have to use various tricks to reduce the sequence lengths based on other induced biases like downsampling rates \citep{clark2022canine,tay2021charformer} or have extremely low training and inference speeds \citep{xue2022byt5}. \citet{chung2016hierarchical} create tokens in a differentiable manner by predicting frontiers and using the representations of each character inside a ``token'', but it remains unclear how their model could be adapted to be used with newer architectures such as Transformers. \citet{mofijul2022vocabulary} propose to segment tokens using a trained ``frontier predictor.'' Nevertheless, this differentiable tokenizer is not trained with the main language model objective but instead mimics a BPE subword tokenizer, carrying some of its flaws.

\section{\protect\ourmodule{}}


\subsection{Differentiable Tokenization}
\label{sec:differentiable_tokenization}
Our main contribution is the introduction of an end-to-end differentiable tokenization architecture that consists in softly aggregating input bytes into what we refer to as \textit{blocks}. As an analogy with hard tokenization schemes, blocks can be compared to tokens with smooth borders.

We decompose the tokenization process into several differentiable operations, ensuring that our model can be trained end-to-end. Our approach consists in predicting a segmentation, and then combining byte embeddings according to this segmentation. \ourmodule{} can be divided in three different parts:
\begin{itemize}
    \item Predicting block frontiers using a parameterized layer to assign a probability $p_{F_i}$ to each input byte $b_i$ of being a frontier;\footnote{$F$ in $p_{F_i}$ stands for \textit{Frontier}.}
    \item Building a byte-block unnormalized joint distribution using the frontier probabilities $(p_{F_i})_{i \in [1, L]}$ corresponding to a soft assignment from bytes to blocks;
    \item Pooling byte representations for each block $B_j$ weighted by the probability of each byte to belong in the current block $P(b_i \in B_j)$.
\end{itemize}

This process results in a sequence of embeddings that can be given directly to the encoder-decoder model. We provide an overview of the entire model in Figure~\ref{fig:overview_diagram}. We also summarize the process of mapping byte embeddings to block embeddings in appendix~\ref{sec:appendix_detailled}.

\subsubsection{Predicting Subword Frontiers}
\label{sec:frontpred}
Our frontier predictor consists in a parameterized module mapping each byte $b_i$ to the probability of being a block frontier $p_{F_i}$. In a first part, we embed each byte $b_i$ to an embedding $e_{b_i}$. Working with bytes instead of characters allows modeling a larger array of symbols while having very small embedding matrices with $256\times hidden\ size$ parameters. Since the input sequences fed to the frontier predictor may be particularly long, we use a Transformer with sliding window attention~\citep{beltagy2020longformer}. This layer achieves a linear complexity with respect to sequence length by computing attention using only a local context. This reduced context forces the model to focus on local surface features rather than long-range dependencies which may be hard to model at the byte level.

We make the assumption that long-range dependencies are not relevant for segmentation and that this reduced context window should not harm the quality of the tokenization.

\subsubsection{Modeling the Byte-Block Assignment}
Once the frontier probabilities $(p_{F_i})_{i \in [1, L]}$ are predicted for the whole sequence, we use them to model an assignment between bytes and block slots. Each byte is given a probability distribution over the available block slots, and the expected block position of a byte in the block sequence increases along the byte sequence (i.e. the next byte is always more likely to be assigned to the next block).

Let us introduce $(B,\ b_i)$, the slot random variables for each byte $b_i$, describing the position of the block containing $b_i$ in the block sequence. In other words, the event $(B=k,\ b_i)$ describes the fact that the $i$-th byte belongs in the $k$-th block. These variables can only take values in $[1, L]$, as there cannot be more blocks than there are bytes. We can model the $(B,\ b_i)$ as a \textit{cumulative sum} of the random variables $F_i$: the position of the block in which a byte belongs is exactly the number of frontier bytes before this one.

Since $F_i \sim \mathcal{B}(p_{F_i})$, we can model the block index variables $B$ depending on the index of the bytes $b$ using the Poisson Binomial distribution $\mathcal{PB}$ which models the cumulative sum of Bernoulli variables: $\left(B,\ b_i\right) \sim \mathcal{PB}\!\left(\left(p_{F_k}\right)_{k \leq i}\right)$. There exists no closed form for this distribution's mass function, but some fast methods have been developed to avoid exploring the whole event tree \cite{BISCARRI201892, poibin_fft}. However, to reduce computational cost, we use a truncated Gaussian kernel $G$ with the same mean and variance to approximate the $(B,\ b_i)$ probability mass function:
$$
\forall k \in [1, L_B], P\!\left(B = k,\ b_i\right) \simeq P_{k,i} \triangleq \frac{1}{Z}G_{\mu_i, \sigma_i}(k)
$$
where $Z=\sum\limits_{1\leq k\leq L_B}G_{\mu_i, \sigma_i}(k)$ is a normalization term, and:
\begin{equation}
  \begin{cases}
    L_B = \min\left(L, \left(\mu_L + 3\sigma_L\right)\right)\\[11pt]
    \mu_i = \sum_{k=1}^{i} p_{F_i} \\[11pt]
    \sigma_i = \sqrt{\sum_{k=1}^{i} p_{F_i}\left(1-p_{F_i}\right)}
  \end{cases}
\end{equation}

\begin{figure}[t]
    \centering\small
    \includegraphics[width=0.85\columnwidth]{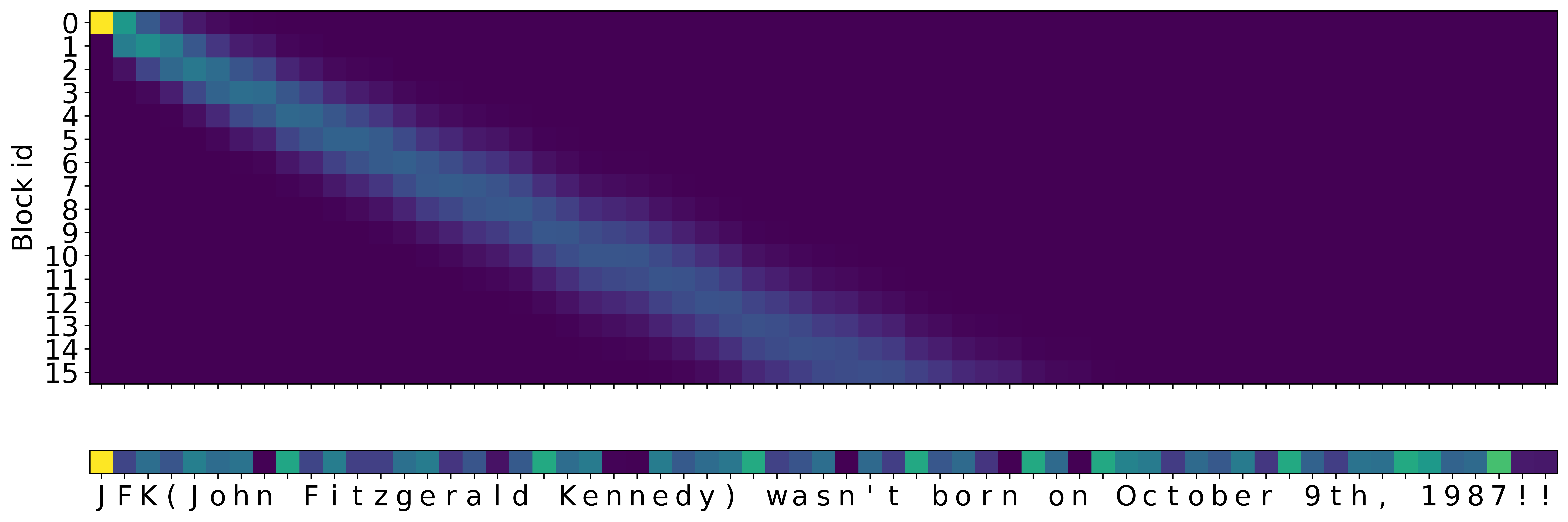}\\
    Step 0\\[2mm]
    \includegraphics[width=0.85\columnwidth]{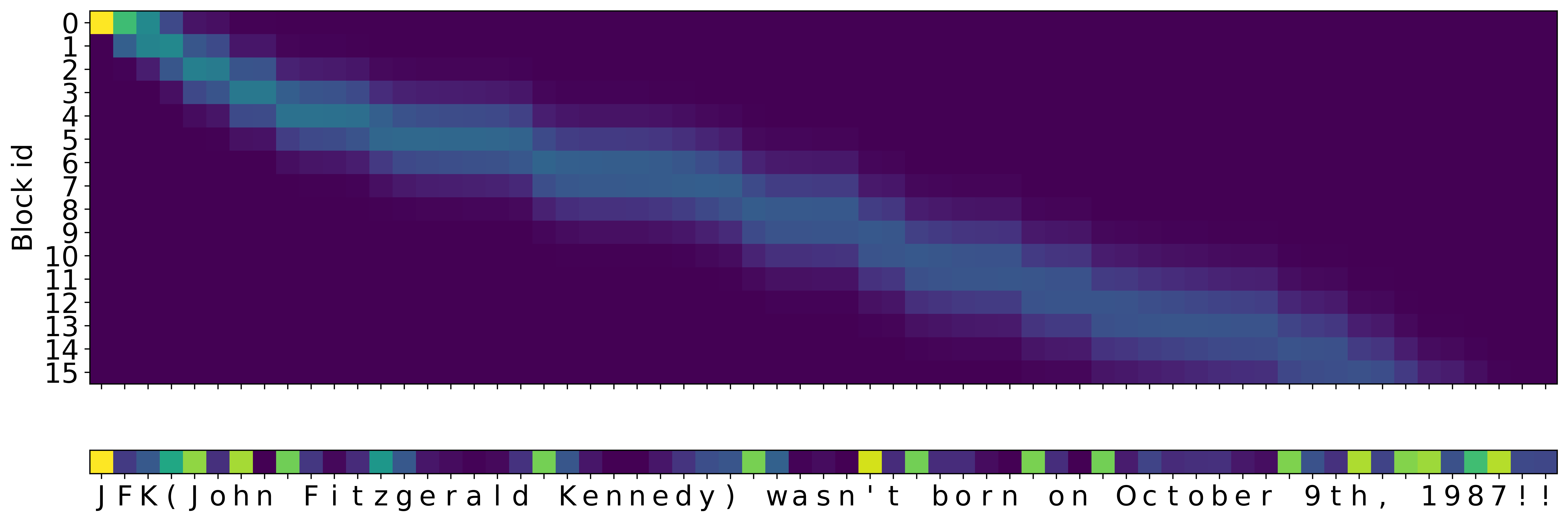}\\
    Step 3,000\\[2mm]
    \includegraphics[width=0.85\columnwidth]{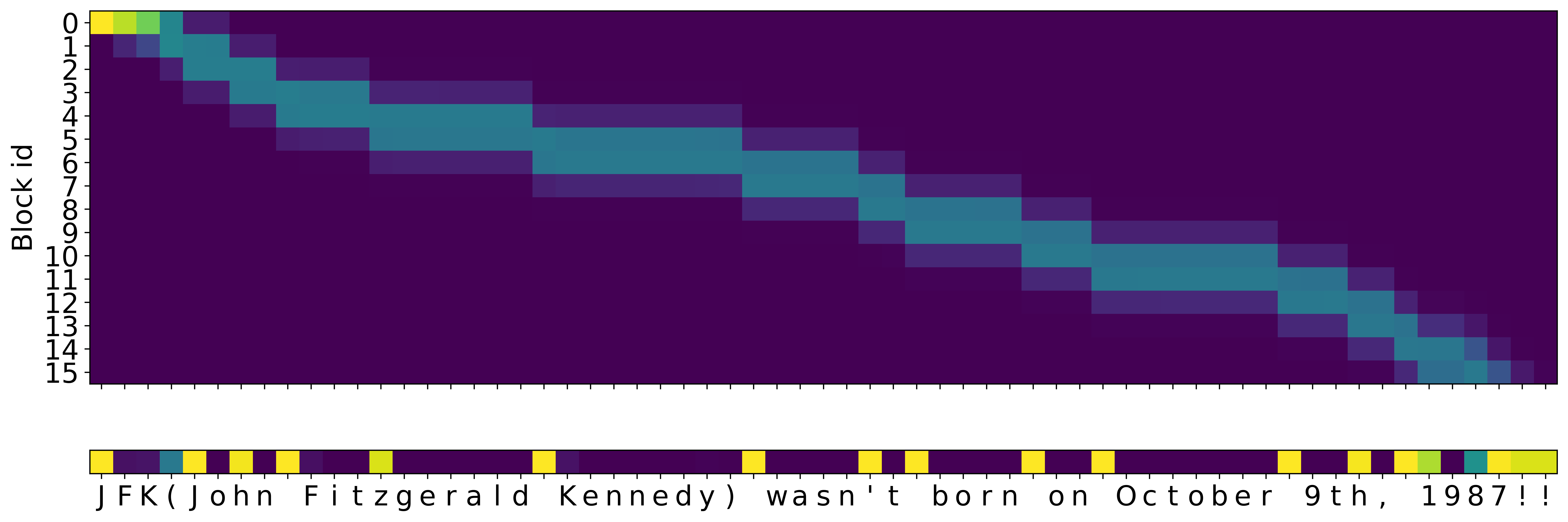}\\
    Step 7,000\\[2mm]
    \includegraphics[width=0.85\columnwidth]{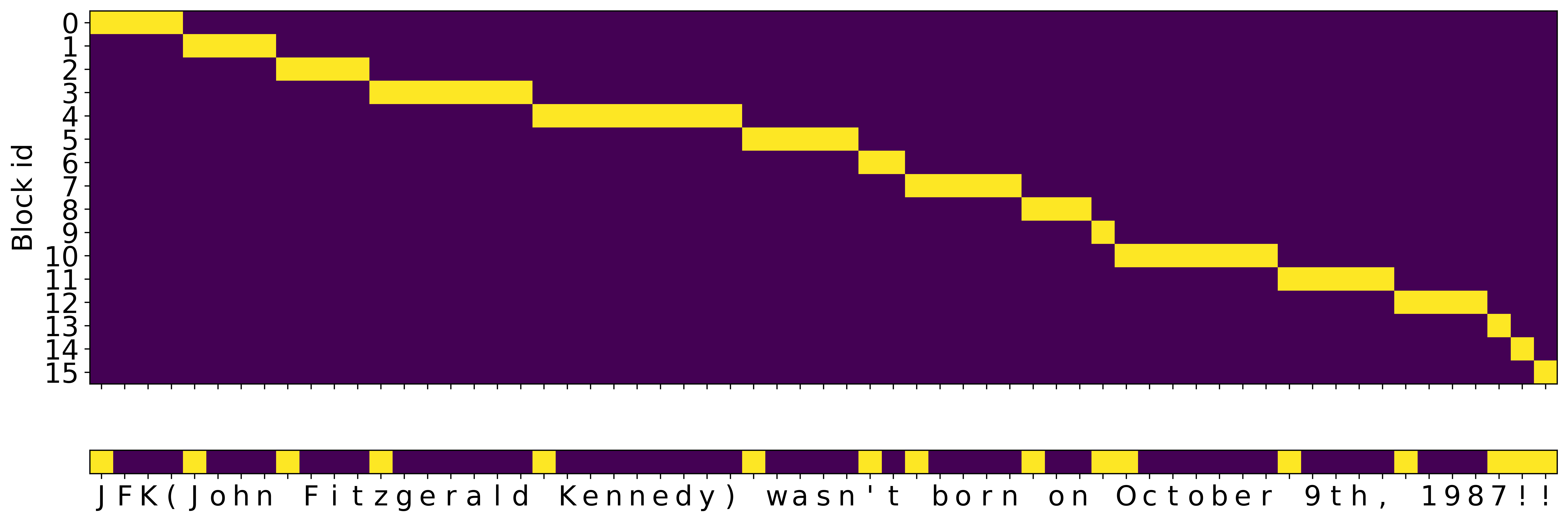}\\
    Step 13,000\\
    \caption{The block-byte assignment $P$ during the first pre-training steps. \ourmodule{} learns to downsample input sequences so that no information is lost through truncation, but also converges towards a sharp segmentation.}
    \label{fig:mapping_example}
\end{figure}

We denote by $P_{k,i}$ the approximation of the probability of membership of the byte $i$ to block $k$. We display an example of this map at different steps during training in Figure~\ref{fig:mapping_example}. We truncate the block sequences after $(\mu_L + 3\sigma_L)$ since all the probabilities beyond this position are negligible.

\subsubsection{Pooling Block Embeddings}
\label{sec:pooling_block}
At this point in the forward pass, we have estimated the position of the block in which each input byte belongs, along with the block sequence maximum plausible length $L_B$.
In order to provide block embeddings to the LM, we now focus on the contribution of each byte to the block given by the block-byte assignment map. For each block position $k \in [1, L_B]$, this map actually provides an unnormalized contribution $(P_{k,i})_{i \in [1, L]}$ of each byte in this block. We can then use the byte embeddings $e_b$ from the frontier predictor described in Section~\ref{sec:frontpred} and, for the $k$-th block, build a block embedding where each byte $b_i$ contributes based on its probability of being in this block $P_{k,i}$.

To build $e^B_k$, the embedding of block $B_k$ in $k\text{-th}$ position, we first compute the weighted byte embeddings ${\left(P_{k,i} \times e^b_i\right)_{i \in [1, L]}}\in \mathbb{R}^H$, with $H$ the hidden size of the byte embeddings. To make the block embeddings aware of the ordering of the bytes (so that \textit{ape} and \textit{pea} can have different representations), we proceed to a depthwise 1-D convolution along the dimension of the bytes after weighting. This convolution also improves the expressiveness of the block embeddings. We discuss our efficient implementation of these operations in Appendix~\ref{sec:appendix_cache}.

We finally apply a max-pooling operation on the contextualized weighted byte embeddings for each block. This yields one embedding per block, with the same dimension as the byte embeddings. We use a linear layer to map the block embeddings to the right input dimension for the encoder-decoder model, i.e. its hidden size.

The final step consists in truncating the block embedding sequence to a fixed length $\hat{L} = \min(L_B, L / K)$ with $K\in \mathbb{N^*}$ a fixed \textit{truncation factor}. This simple heuristic ensures that all sequences fed to the encoder-decoder have a length at least $K$ times shorter than the input byte sequence length. We choose $K=4$ throughout the paper which is in average the number of bytes in an English BPE token. Most importantly, this truncation incentivizes the frontier predictor to produce sufficiently long blocks. We discuss the influence of this mechanism in more depth in Section~\ref{sec:truncation}.


\subsection{Model Training}

We obtain from the differentiable tokenizer and pooling module a sequence of block embeddings that can be used exactly like subword embeddings. Thus, we use an encoder decoder architecture identical to T5~\citep{raffel2020t5}. Nevertheless, since we do not have a fixed subword vocabulary, our decoder operates at the byte level similarly to ByT5~\citep{xue2022byt5}.



\begin{table*}[]
\centering\small
\begin{tabular}{cccccccccc}
\toprule
Model                                   & $|\theta|$ & MNLI      & QNLI & MRPC      & SST-2 & QQP       & STSB & COLA & AVG  \\ \midrule
$\text{T5}_{Small}$                     & 60M       & \textbf{79.7/79.7} & \textbf{85.7} & 80.2/86.2 & 89.0  & \textbf{90.2/86.6} & 80.0 & 30.3 & 76.6 \\[3pt]
$\text{\ourmodel{}}_{Small}$ (ours)     & 57M          & 79.2/78.6 & 84.5 & \textbf{82.3/87.2}    & \textbf{89.6}  & 89.9/\textbf{86.5}    & \textbf{81.4} & \textbf{32.0} & \textbf{77.1} \\ \bottomrule
\end{tabular}
\caption{Results on dev sets for the GLUE benchmark for small models following our pre-training procedure.}
\label{tab:glue_small}
\end{table*}
\begin{table*}[t]
\centering\small
\begin{tabular}{cccccccccc}
\toprule
Model                                   & $|\theta|$ & MNLI      & QNLI & MRPC      & SST-2 & QQP       & STSB & COLA & AVG  \\ \midrule
$\text{BERT}_{Base}^\dagger$            & 110M       & \textbf{84.4} / -  & 88.4 & 86.7/-    & \textbf{92.7}  & -         & -    & -    & -    \\[3pt]
$\text{T5}_{Base}^\dagger$              & 220M       & 84.2/\textbf{84.6} & 90.5 & \textbf{88.9/92.1} & \textbf{92.7} & \textbf{91.6/88.7} & \textbf{88.0} & 53.8 & 84.3 \\ \midrule
$\text{CharBERT}_{Base}^\mathsection$   & 125M       & -         & \textbf{91.7} & 87.8/-    & -     & 91/-      & -    & \textbf{59.1} & -    \\[3pt]
$\text{Byte-level T5}_{Base}^\dagger$   & 200M       & 82.5/82.7 & 88.7 & 87.3/91.0 & 91.6  & 90.9/87.7 & 84.3 & 45.1 & 81.5 \\[3pt]
$\text{Charformer}_{Base}^\dagger$      & 203M       & 82.6/82.7 & 89.0 & 87.3/91.1 & 91.6  & 91.2/88.1 & 85.3 & 42.6 & 81.4 \\[3pt]
$\text{\ourmodel{}}_{Base}$ (ours)     & 200M        & 77.5/78.8 & 88.2 & 82.4/88.2 & 91.3  & 90.8/87.7 & 79.2 & 51.0 & 80.3\\ [3pt] \bottomrule
\end{tabular}
\caption{Results on dev sets for the GLUE benchmark. $\dagger$ indicates results obtained by \citet{tay2021charformer}, which are very similar to our models in terms of compute, but use a smaller batch size which may enhance their performance. $\mathsection$ indicates results obtained by \citet{ma-etal-2020-charbert}. The top section concerns model trained using a subword tokenizer.}
\label{tab:glue}
\end{table*}

\subsubsection{Pre-Training Details}

\paragraph{Objective}
Our objective is identical to the one used in ByT5. We mask 15\% of bytes randomly and choose a number of spans such that each has an average length of 20 bytes. Each span is then replaced by an \texttt{<extra\_id\_i>} token with \texttt{i} identifying the order of the span in the sequence. On the decoder side, the model has to predict in an autoregressive way the span identifier and the masked bytes.

\paragraph{Data}
We pre-train our model on English text data using C4~\cite{raffel2020t5}, a large corpus scraped from the Internet. This corpus is particularly suited to our pre-training due to its diversity in terms of content and linguistic variations. In addition, it enables a better comparison with other tokenizer-free models trained using it such as Charformer. Since this dataset is not available publicly, we use the English split of the mC4 distributed by AllenAI. We filter long documents containing more than $2^{15}$ bytes, which is a simple proxy to remove important quantities of unwanted code data.

\paragraph{Hyperparameters}
We pre-train two versions of our model: $\text{\ourmodel{}}_{Small}$ and $\text{\ourmodel{}}_{Base}$. Each of them stacks a $\text{\ourmodule{}}_{Small}$ (resp. $\text{\ourmodule{}}_{Base}$) tokenizer and embedding module and a $\text{T5}_{Small}$ (resp. $\text{T5}_{Base}$) encoder-decoder model stripped of its tokenizer and subword embedding matrix. Details about $\text{\ourmodule{}}$ hyperparameters can be found in Appendix~\ref{sec:appendix_hp}.

Following T5 and ByT5, we use the Adafactor optimizer with a learning rate of $10^{-2}$ for the encoder-decoder model, parameter scaling for the whole system and no weight decay. However, to maintain stability of our differentiable tokenizer, we use a learning rate of $10^{-3}$ for the parameters of the byte embeddings, the frontier predictor, and the pooling module. We also use a triangular learning rate schedule with 1000 (resp. 5000) warm-up steps for batch size 1024 (resp. 64).

\paragraph{Training} We train $\text{T5}_{Small}$, $\text{\ourmodel{}}_{Small}$, and $\text{\ourmodel{}}_{Base}$ for 65k steps with a batch size of 1024. Sequence lengths are respectively 1024 for $Small$ models and 2048 for the $Base$ model. Thus, the models are trained on roughly the same amount of bytes as in \citet{tay2021charformer}, where a batch size of 64 is used for 1M steps. 

We also train a $\text{ByT5}_{Small}$ model on the same data, using a batch size of 64 and a sequence length of 1024. We consider the ``Scaled'' architecture which provides the encoder with more layers than the decoder \cite{xue2022byt5}. To avoid prohibitive computation costs and ensure fairness in terms of available resources between models, we limit its training time to the one of $\text{\ourmodel{}}_{Small}$. Hence, our $\text{ByT5}_{Small}$ is only trained for 200k steps.

\section{Experiments and Results}
\subsection{Evaluation on GLUE}

To ensure that our model is competitive with existing language models exploiting subword tokenization algorithms, we evaluate it on several English datasets and compare it with other baseline models.

\paragraph{Setup} We use GLUE~\cite{wang-etal-2018-glue}, a Natural Language Understanding benchmark consisting of 7 tasks, to evaluate our model. Similarly to T5, we cast the classification tasks as generation tasks where the model has to predict autoregressively the bytes forming the answer. 

We compare our model to an encoder-decoder model with subword tokenization (pre-trained with the same denoising objective as T5) and a fully byte-level encoder-decoder, similar to ByT5. We compare $Small$ models with our pre-trained versions, and $Base$ models with results mentioned in \citet{tay2021charformer}. We report the number of parameters given in \citet{tay2021charformer} for $\text{Byte-level T5}_{Base}$, and gather from its low value that their implementation corresponds to a $\text{T5}_{Base}$ architecture trained on byte-level inputs. 

\paragraph{Results} Results can be found on Tables~\ref{tab:glue_small} and~\ref{tab:glue}. Overall, \ourmodel{} exhibits a performance slightly below Charformer but stays within a small margin on average (1.1 points below). Nonetheless, the main objective of our method is to balance decent performance with robustness and speed which we show in the following sections.


\subsection{Robustness to Domain Change}
Static subword tokenizers tend to show important limitations when used with texts originating from a domain unseen during training. For instance, \citet{el-boukkouri-etal-2020-characterbert} show that tokenizing medical texts with a tokenizer trained on Wikipedia data often results in an over-segmentation of technical terms which in turn affects the downstream performance. By removing this static bottleneck in \ourmodel{}, we hope that it should be able to adapt more easily to new domains. To test this hypothesis, we finetune it on a medical Natural Language Inference dataset.

\begin{table}[t]
\centering\small
\begin{tabular}{lc}
\toprule
Model                                   & Accuracy  \\ \midrule
$\text{BERT}_{Base}^\ddagger$           & 77.7      \\
$\text{CharacterBERT}_{Base}^\ddagger$  & 77.9      \\
$\text{T5}_{Small}$                     & 75.3      \\ 
$\text{\ourmodel{}}_{Small}$ (ours)     & 75.6      \\\bottomrule
\end{tabular}
\caption{Results on MedNLI. $\ddagger$ indicates results from \citet{el-boukkouri-etal-2020-characterbert}, who use a different pre-training corpus than C4. All other results are from models trained with our codebase.}
\label{tab:mednli}
\end{table}

\paragraph{Setup} We finetune \ourmodel{} on \textsc{MedNLI}~\cite{romanov-shivade-2018-lessons}, a dataset consisting of 14,049 sentence pairs extracted from clinical notes. We follow the same finetuning setup than for the GLUE Benchmark i.e. use the same batch size and learning rate. We compare our results to the ones obtained by~\citet{el-boukkouri-etal-2020-characterbert} with models pretrained on the general domain.

\paragraph{Results} We present our results on Table~\ref{tab:mednli}. Although we notice a significant drop in performance compared to the encoder models trained by~\cite{el-boukkouri-etal-2020-characterbert}, we believe this drop may be due to the different pretraining data used---CharacterBERT uses splits of Wikipedia, which may be helpful to learn some technical terms related to the clinical domain---, and the different model sizes---CharacterBERT uses all of its parameters to encode example, while we keep half of the parameters in the decoder. Nonetheless, we note that \ourmodel{} reaches a better performance than its subword tokenization counterpart T5.

\subsection{Robustness to Noisy Data}
Although LMs may learn complex patterns even from noisy input texts, this ability is conditioned by how the tokenizer segments character sequences. Since \ourmodule{} is not static and can be finetuned on non-standard data, we expect it should be able to learn to be more robust to variation/noise compared to a subword tokenizer paired with a LM. To evaluate this hypothesis, we study how \ourmodel{} behaves on both naturally occurring text variation and multiple levels of synthetic noise.

\subsubsection{Naturally Occurring Noise}
\paragraph{Setup} Similarly to \citet{tay2021charformer}, we test our model on a toxicity detection task constructed with user generated data. We use the \textsc{ToxicComments}{} dataset~\cite{wulczyn2017ex} which contains 223,549 sentences annotated with a binary label indicating whether each sentence can be classified as toxic or not. We also use the same finetuning setup here as the one used for evaluating on the GLUE benchmark. 

\paragraph{Results} We present our results in Table~\ref{tab:toxic} and compare them to the ones reported in \citet{tay2021charformer}. As expected, noisy user generated data is particularly harmful for models using subword tokenization. On the other hand, constructing sentence representations with byte-level information  helps and our model is more accurate than Charformer. This gain may be due to a better segmentation of specific terms encountered in the data.

\subsubsection{Synthetic Noise}
\paragraph{Setup} We also compare T5 and ByT5 with our approach when facing different levels of noise. This study pictures how these models react to unseen noise at evaluation time (\textsc{Dev-Only} setup) and how they adapt to a given noise via fine-tuning (\textsc{Train-Dev} setup). We apply synthetic noise at different levels $\tau\in\{0.05, 0.10, 0.15\}$ by picking randomly $\tau \times L$ positions in the byte sequences and equiprobably deleting, replacing or inserting bytes at these positions.

\begin{table}[t]
\centering\small
\begin{tabular}{lc}
\toprule
Model                                   & Accuracy      \\ \midrule
$\text{T5}_{Base}^\dagger$              & 91.5          \\
$\text{Charformer}_{Base}^\dagger$      & 92.7          \\
$\text{\ourmodel{}}_{Base}$ (ours)      & {\bf 93.2}    \\\bottomrule
\end{tabular}
\caption{Results on the \textsc{ToxicComments} dataset. Results indicated by $\dagger$ are from~\citet{tay2021charformer}.}
\label{tab:toxic}
\end{table}

\begin{figure}[t]
\centering\small
\begin{subfigure}[b]{\columnwidth}
\centering\small
\includegraphics[width=0.9\columnwidth]{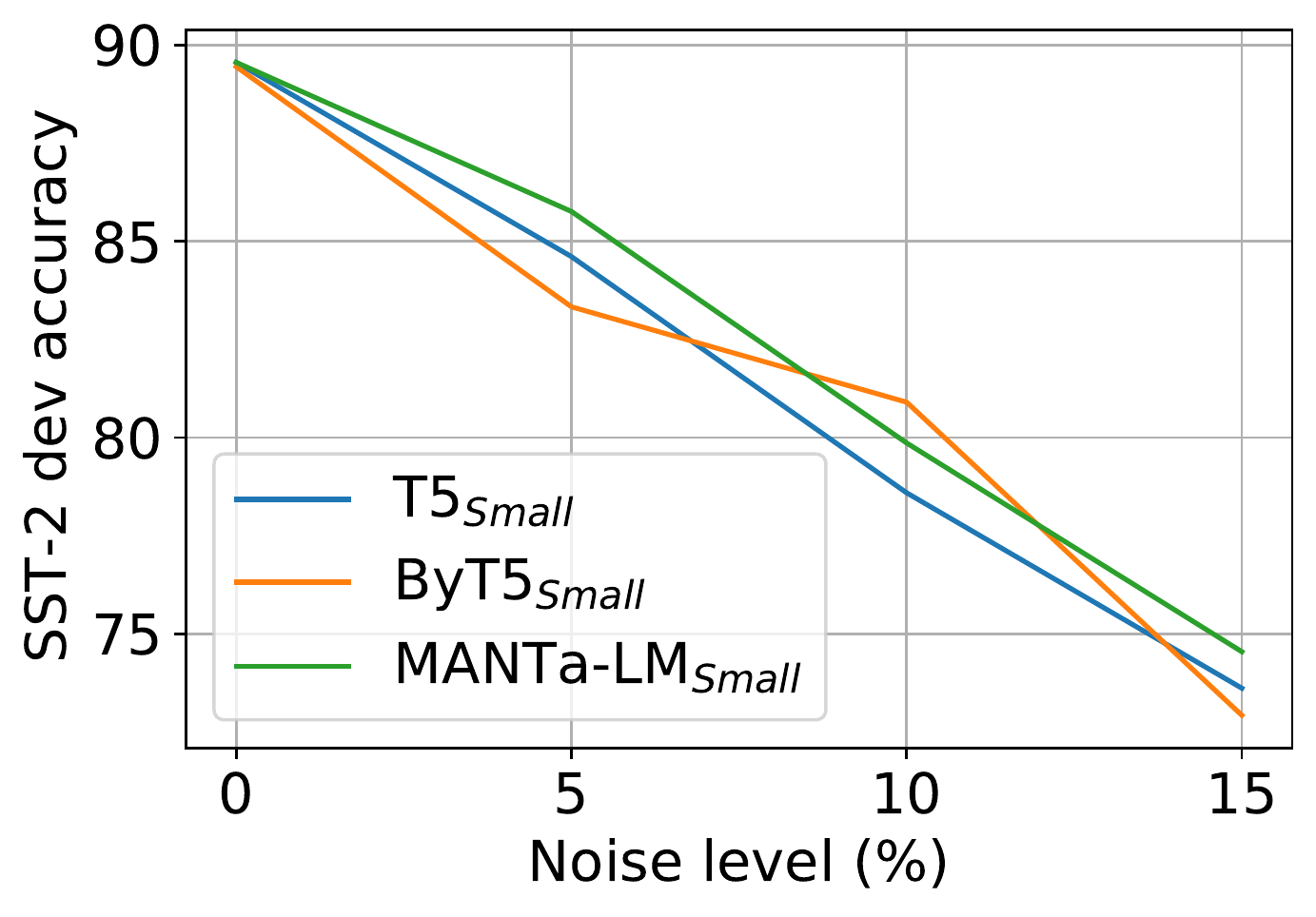}\vspace*{-2mm}
\caption{\textsc{Dev-Only}}
\end{subfigure}

\vspace{2mm}

\begin{subfigure}[b]{\columnwidth}
\centering\small
\includegraphics[width=0.9\columnwidth]{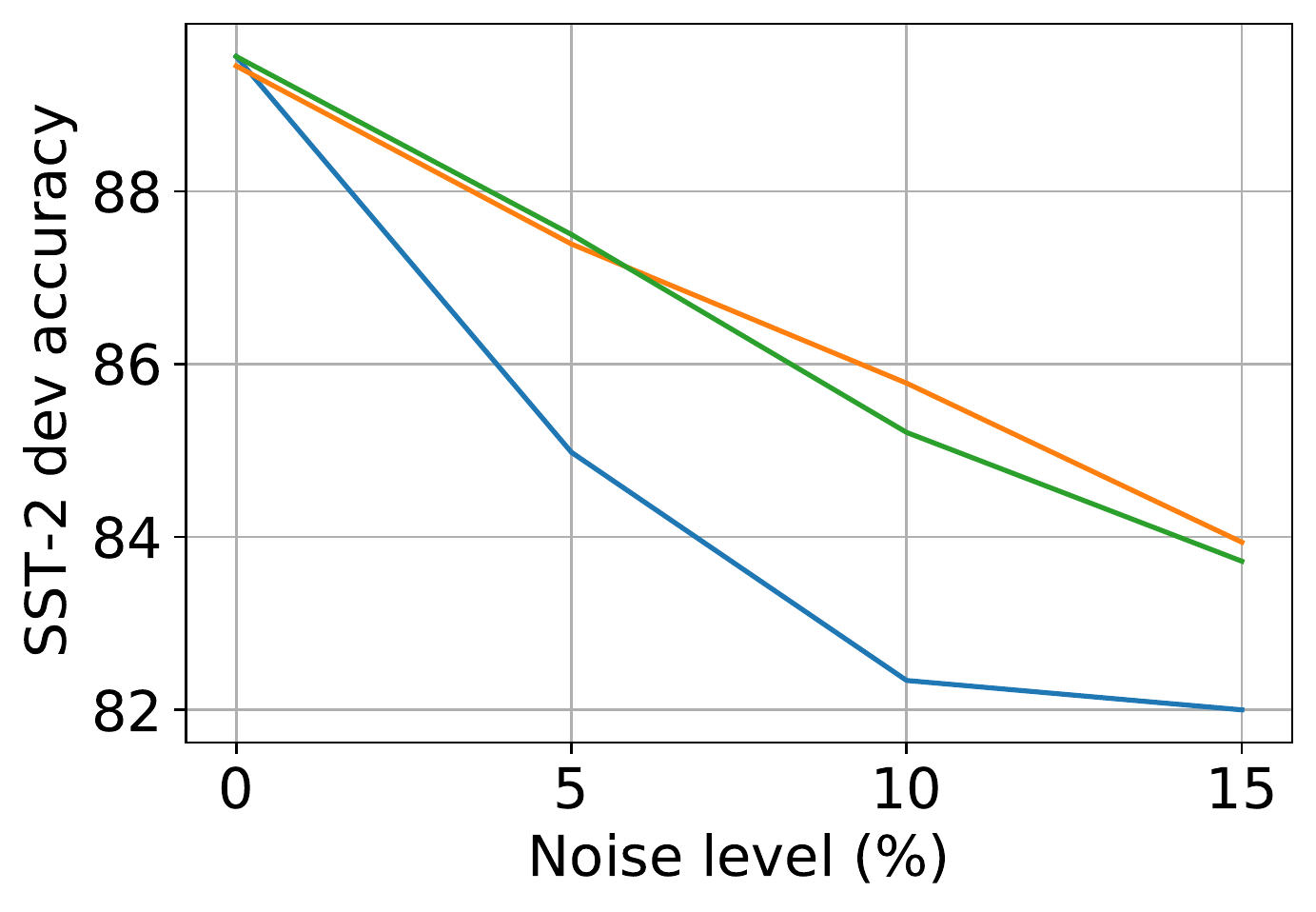}\vspace*{-2mm}
\caption{\textsc{Train-Dev}}
\end{subfigure}
\caption{Best accuracy on the SST-2 development set as the noise level increases. The \textsc{Train-Dev} setting corresponds to models finetuned on noisy data while models in the \textsc{Dev-Only} setting have been finetuned on clean data.}
\end{figure}


\paragraph{Results} We found that models performed similarly for the different noise levels in the \textsc{Dev-Only} setting. On the contrary, in the \textsc{Train-Dev} setting, \ourmodel{} can be finetuned as well as ByT5 for all levels of noise, while the performance of T5 quickly degrades.

\begin{table}[t]
\centering\small
\begin{tabular}{lrr}
\toprule
Model                           & $|\theta|$ & Seconds/step         \\ \midrule
$\text{Byte-level T5}_{Small}$   & 57M       & 9.06 ($\times$ 8.0)  \\[3pt]
$\text{\ourmodel{}}_{Small}$    & 57M        & 2.61 ($\times$ 2.3)    \\[3pt]
$\text{T5}_{Small}$      & 60M        & 1.13 ($\times$ 1)      \\ \bottomrule
\end{tabular}
\caption{Comparison of training speeds. All the experiments were run on 16 NVIDIA V100 GPUs using a batch size of 1024 and a sequence length of 1024 bytes or 256 tokens}
\label{tab:speed}
\end{table}

\section{Training Speedups}

In terms of speed, we compare our model to $\text{\ourmodel{}}_{Small}$ to $\text{T5}_{Small}$ counterparts: one that is trained at the classical subword-level, and one trained at byte-level, hence using sequences that are roughly 4 times longer. We also report the speed of the larger $\text{ByT5}_{Small}$ architecture as described in \citet{xue2022byt5}.

\ourmodel{} is approximately 4 times faster than $\text{Byte-level T5}_{Small}$, and 5 times faster than $\text{ByT5}_{Small}$, which can be explained by the reduced sequence length we use in the encoder-decoder model. \ourmodel{} is only 2.3 times slower than $\text{T5}_{Small}$ which furthermore benefits from already tokenized sequences at training time.

\section{Discussion}
\label{sec:discussion}

\begin{table*}[t]
\centering\small
\begin{tabular}{lc}
\toprule
\textbf{Original}                & Oh, it's me vandalising?xD See here. Greetings,         \\
\textbf{\ourmodule{}}            & \texttt{O\nors{}h\rs{},\rs{} \nors{}i\nors{}t\rs{}'\nors{}s\rs{} \nors{}m\nors{}e\rs{} \nors{}v\nors{}a\nors{}n\nors{}d\nors{}a\nors{}l\nors{}i\nors{}s\nors{}i\nors{}n\nors{}g\nors{}?\rs{}x\nors{}D\rs{} \nors{}S\nors{}e\nors{}e\rs{} \nors{}h\nors{}e\nors{}r\nors{}e\rs{}.\rs{} \nors{}G\nors{}r\nors{}e\nors{}e\nors{}t\nors{}i\nors{}n\nors{}g\nors{}s\rs{},}         \\
\textbf{T5 tokenizer}            & \texttt{O\nors{}h\rs{},\rs{} \nors{}i\nors{}t\rs{}'\rs{}s\rs{} \nors{}m\nors{}e\rs{} \nors{}v\nors{}a\nors{}n\rs{}d\rs{}a\nors{}l\rs{}i\nors{}s\nors{}i\nors{}n\nors{}g\rs{}?\rs{}x\rs{}D\rs{} \nors{}S\nors{}e\nors{}e\rs{} \nors{}h\nors{}e\nors{}r\nors{}e\rs{}.\rs{} \rs{}G\nors{}r\nors{}e\nors{}e\nors{}t\nors{}i\nors{}n\nors{}g\rs{}s\rs{},}\\ \midrule 

\textbf{Original}                & The patient was started on Levophed at 0.01mcg/kg/min. \\
\textbf{\ourmodule{}}            & \texttt{T\nors{}h\nors{}e\rs{} \nors{}p\nors{}a\nors{}t\nors{}i\nors{}e\nors{}n\nors{}t\rs{} \nors{}w\nors{}a\nors{}s\rs{} \nors{}s\nors{}t\nors{}a\nors{}r\nors{}t\nors{}e\nors{}d\rs{} \nors{}o\nors{}n\rs{} \nors{}L\nors{}e\nors{}v\nors{}o\nors{}p\nors{}h\nors{}e\nors{}d\rs{} \nors{}a\nors{}t\rs{} \nors{}0\rs{}.\nors{}0\nors{}1\nors{}m\nors{}c\nors{}g\rs{}/\nors{}k\nors{}g\rs{}/\nors{}m\nors{}i\nors{}n\rs{}.} \\
\textbf{T5 tokenizer}            & \texttt{T\nors{}h\nors{}e\rs{} \nors{}p\nors{}a\nors{}t\nors{}i\nors{}e\nors{}n\nors{}t\rs{} \nors{}w\nors{}a\nors{}s\rs{} \rs{}s\nors{}t\nors{}a\nors{}r\nors{}t\nors{}e\nors{}d\rs{} \nors{}o\nors{}n\rs{} \nors{}L\nors{}e\rs{}v\nors{}o\rs{}p\rs{}h\nors{}e\rs{}d\rs{} \nors{}a\nors{}t\rs{} \nors{}0\nors{}.\rs{}0\nors{}1\rs{}m\rs{}c\rs{}g\rs{}/\rs{}k\nors{}g\rs{}/\rs{}m\nors{}i\nors{}n\rs{}.} \\\bottomrule
\end{tabular}
\caption{Examples of segmentations produced by our module (pre-trained only) and by T5's BPE tokenizer. The sentences are samples from \textsc{ToxicComments} and \textsc{MedNLI}.}
\label{tab:segmentation}
\end{table*}

\subsection{Truncating Embedding Sequences}
\label{sec:truncation}
Once we obtain block embeddings, the final step in \ourmodule{} consists in truncating sequences to a length 4 times smaller than the original byte sequence, as described in Section~\ref{sec:pooling_block}. This is essential to make \ourmodel{} work.

First, it increases the control over the encoder-decoder's computation cost. Without this bottleneck, the Transformer can receive sequences varying from a single block containing the whole sequence ($L_B=1$) to one block per byte in the sequence ($L_B = L$). In the latter case, which mimics ByT5's input segmentation, the computation becomes extremely slow due to the quadratic cost  of the attention with respect to the sequence length. Using the bottleneck ensures that we can control the worst case complexity of the encoder Transformer and keep it similar to that of a subword-based encoder model.

Second, it serves as a kind of regularization for the block segmentations. We noted that training our module without the bottleneck often led to block sequences as long as byte sequences ($L_B=L$). This may be due to the beginning of training where having very local information helps - for instance bytes to the left and right of masked spans. However, such a segmentation degrades the model speed and performance later in training. Truncating the sequence forces the model to construct larger blocks in order to ``fit'' all the information from the input sequence.

\subsection{Learnt Block Segmentation}
Segmentation examples can be found in Table \ref{tab:segmentation}. For each byte, we retrieve the expected block position produced by \ourmodule{} and approximate it with the closest integer to mimic hard tokenization. We found that \ourmodule{} is not keen to produce subword level segmentations. Most of the key symbols for word separation have been identified as block delimiters during pre-training. As expected, \ourmodule{} is less prone to over-segmentation of unknown words like named entities. We also found that a trained \ourmodule{} produced spiked separation probabilities, meaning that it converged towards a ``hard'' segmentation. This can also be observed by monitoring the value $\min(p_{F_i}, 1 - p_{F_i})$ which always converges towards values of magnitude $10^{-5}$.

\subsection{Gradient-Based Segmentation}

We employ a radically different downsampling approach compared to other gradient-based tokenization methods such as CANINE~\cite{clark2022canine} or Charformer~\cite{tay2021charformer}. While CANINE downsamples sequences using a fixed rate after byte contextualization and Charformer's GBST (Gradient Based Subword Tokenizer) pools representations created using various downsampling rates, \ourmodule{} only applies downsampling right before the LM to limit the length of block sequences. Hence, our model is able to build word-level representations of \textit{arbitrary length} as long as it divides the whole byte sequence length by a fixed factor.

We also argue that our method yields more explainable pooled representations as the segmentation can be explicitly derived from the outputs of \ourmodule{}. Indeed, contrary to CANINE and Charformer, \ourmodule{} disentangles the segmentation of blocks from their representations, allowing to study each part separately.

\subsection{Main hyperparameters}
We discuss here some of the major hyperparameters of our method. Constrained by limited computational resources, we were unable to assess their exact importance on MANTa's performance. We try to give some intuitions on their influence.

\paragraph{Frontier Predictor} We used a small Transformer network with sliding window attention for this module. A much larger network would be slower and may not bring significant improvements to the overall performance of the model, since it is only used for predicting the block byte assignment but does not ``expand'' the overall expressivity of the model.

\paragraph{Convolution kernel applied on byte embeddings} This kernel adds positional information to the byte embeddings and expressivity when constructing the block embeddings. Using a larger kernel or a concatenation of kernels might help for better block representations. However, our experiments did not show any significant difference in the pretraining performance.

\paragraph{Block embedding sequence truncation factor} Trimming block sequences was instrumental to produce meaningful input segmentations and blocks containing more than a single byte. We settled for a factor of 4 since other values led to minor degradations early in training. This factor roughly corresponds to the average number of bytes in a subword created by an English tokenizer.

We believe that a more thorough hyperparameter search could improve the performance of our model. We leave this for future work due to computational limitations.

\section{Conclusion}
In this work, we present \ourmodule{}, a fully differentiable module that learns to segment input byte sequences into blocks of arbitrary lengths, and constructs a robust representation for these blocks. We train this module jointly with an encoder-decoder LM on a span denoising objective to obtain \ourmodel{}. We then show that \ourmodel{} is more robust when applied to noisy or out-of-domain data than models using static subword tokenizers. At the same time, it performs on par with fully byte-level models on these setups while operating with a much reduced computational cost.

Beyond the noisy and out-of-domain settings, we believe that our approach could lead to interesting results for a number of languages, especially those whose writing system do not use whitespace separators, such as Chinese.

Finally, tokenizers are hypothesized to be an important limiting factor when segmenting multilingual data~\cite{rust-etal-2021-good}. We believe \ourmodule{} could be used in the multilingual setting to ensure a more balanced segmentation between languages.

\section*{Limitations}
Although \ourmodule{} can help alleviate some of the inherent issues accompanying subword tokenizers, it also suffers some flaws that we believe could be addressed in future work.

Contrary to encoder-decoder models that can decode long sequences efficiently, our model has to decode sequences byte-per-byte (similarly to~\citet{clark2022canine,xue2022byt5,tay2021charformer}) which adds an important computational overhead at generation time. Previous works have attempted to reduce this computational cost by decreasing the size of the decoder layers compared to the encoder~\cite{xue2022byt5} or by projecting embeddings to a smaller latent space~\cite{jaegle2021perceiver} for the decoding.

Finally, we presented in this work a proof of concept of adaptive segmentation algorithms on relatively small models, ranging from 50M to 200M parameters. Although we hypothesize that our model would scale relatively well since it keeps most of the encoder-decoder architecture untouched, this hypothesis should be tested in a future work.

\section*{Acknowledgements}

This work was funded by the last authors' chair in the PRAIRIE institute funded by the French national agency ANR as part of the ``Investissements d'avenir'' programme under the reference ANR-19-P3IA-0001.
This work was granted access by GENCI to the HPC resources of IDRIS under the allocation 2022-AD011012676R1. 

\bibliography{anthology,custom}

\begin{thebibliography}{29}
\expandafter\ifx\csname natexlab\endcsname\relax\def\natexlab#1{#1}\fi

\bibitem[{Beltagy et~al.(2020)Beltagy, Peters, and
  Cohan}]{beltagy2020longformer}
Iz~Beltagy, Matthew~E Peters, and Arman Cohan. 2020.
\newblock Longformer: The long-document transformer.
\newblock \emph{arXiv preprint arXiv:2004.05150}.

\bibitem[{Biscarri et~al.(2018)Biscarri, Zhao, and Brunner}]{BISCARRI201892}
William Biscarri, Sihai~Dave Zhao, and Robert~J. Brunner. 2018.
\newblock \href {https://doi.org/https://doi.org/10.1016/j.csda.2018.01.007} {A
  simple and fast method for computing the poisson binomial distribution
  function}.
\newblock \emph{Computational Statistics \& Data Analysis}, 122:92--100.

\bibitem[{Bostrom and Durrett(2020)}]{bostrom-durrett-2020-byte}
Kaj Bostrom and Greg Durrett. 2020.
\newblock \href {https://doi.org/10.18653/v1/2020.findings-emnlp.414} {Byte
  pair encoding is suboptimal for language model pretraining}.
\newblock In \emph{Findings of the Association for Computational Linguistics:
  EMNLP 2020}, pages 4617--4624, Online. Association for Computational
  Linguistics.

\bibitem[{Chung et~al.(2016)Chung, Ahn, and Bengio}]{chung2016hierarchical}
Junyoung Chung, Sungjin Ahn, and Yoshua Bengio. 2016.
\newblock Hierarchical multiscale recurrent neural networks.
\newblock \emph{arXiv preprint arXiv:1609.01704}.

\bibitem[{Clark et~al.(2022)Clark, Garrette, Turc, and
  Wieting}]{clark2022canine}
Jonathan~H Clark, Dan Garrette, Iulia Turc, and John Wieting. 2022.
\newblock Canine: Pre-training an efficient tokenization-free encoder for
  language representation.
\newblock \emph{Transactions of the Association for Computational Linguistics},
  10:73--91.

\bibitem[{Devlin et~al.(2019)Devlin, Chang, Lee, and
  Toutanova}]{devlin-etal-2019-bert}
Jacob Devlin, Ming-Wei Chang, Kenton Lee, and Kristina Toutanova. 2019.
\newblock \href {https://doi.org/10.18653/v1/N19-1423} {{BERT}: Pre-training of
  deep bidirectional transformers for language understanding}.
\newblock In \emph{Proceedings of the 2019 Conference of the North {A}merican
  Chapter of the Association for Computational Linguistics: Human Language
  Technologies, Volume 1 (Long and Short Papers)}, pages 4171--4186,
  Minneapolis, Minnesota. Association for Computational Linguistics.

\bibitem[{El~Boukkouri et~al.(2020)El~Boukkouri, Ferret, Lavergne, Noji,
  Zweigenbaum, and Tsujii}]{el-boukkouri-etal-2020-characterbert}
Hicham El~Boukkouri, Olivier Ferret, Thomas Lavergne, Hiroshi Noji, Pierre
  Zweigenbaum, and Jun{'}ichi Tsujii. 2020.
\newblock \href {https://doi.org/10.18653/v1/2020.coling-main.609}
  {{C}haracter{BERT}: Reconciling {ELM}o and {BERT} for word-level
  open-vocabulary representations from characters}.
\newblock In \emph{Proceedings of the 28th International Conference on
  Computational Linguistics}, pages 6903--6915, Barcelona, Spain (Online).
  International Committee on Computational Linguistics.

\bibitem[{Garcia et~al.(2021)Garcia, Constant, Parikh, and
  Firat}]{garcia-etal-2021-towards}
Xavier Garcia, Noah Constant, Ankur Parikh, and Orhan Firat. 2021.
\newblock \href {https://doi.org/10.18653/v1/2021.naacl-main.93} {Towards
  continual learning for multilingual machine translation via vocabulary
  substitution}.
\newblock In \emph{Proceedings of the 2021 Conference of the North American
  Chapter of the Association for Computational Linguistics: Human Language
  Technologies}, pages 1184--1192, Online. Association for Computational
  Linguistics.

\bibitem[{Graves(2013)}]{graves2013generating}
Alex Graves. 2013.
\newblock Generating sequences with recurrent neural networks.
\newblock \emph{arXiv preprint arXiv:1308.0850}.

\bibitem[{Jaegle et~al.(2021)Jaegle, Borgeaud, Alayrac, Doersch, Ionescu, Ding,
  Koppula, Zoran, Brock, Shelhamer, H{\'{e}}naff, Botvinick, Zisserman,
  Vinyals, and Carreira}]{jaegle2021perceiver}
Andrew Jaegle, Sebastian Borgeaud, Jean{-}Baptiste Alayrac, Carl Doersch,
  Catalin Ionescu, David Ding, Skanda Koppula, Daniel Zoran, Andrew Brock, Evan
  Shelhamer, Olivier~J. H{\'{e}}naff, Matthew~M. Botvinick, Andrew Zisserman,
  Oriol Vinyals, and Jo{\~{a}}o Carreira. 2021.
\newblock \href {http://arxiv.org/abs/2107.14795} {Perceiver {IO:} {A} general
  architecture for structured inputs {\&} outputs}.
\newblock \emph{CoRR}, abs/2107.14795.

\bibitem[{J{\'o}zefowicz et~al.(2016)J{\'o}zefowicz, Vinyals, Schuster,
  Shazeer, and Wu}]{Jzefowicz2016ExploringTL}
Rafal J{\'o}zefowicz, Oriol Vinyals, Mike Schuster, Noam~M. Shazeer, and
  Yonghui Wu. 2016.
\newblock Exploring the limits of language modeling.
\newblock \emph{ArXiv}, abs/1602.02410.

\bibitem[{Kim et~al.(2016)Kim, Jernite, Sontag, and Rush}]{kim2016character}
Yoon Kim, Yacine Jernite, David Sontag, and Alexander~M Rush. 2016.
\newblock Character-aware neural language models.
\newblock In \emph{Thirtieth AAAI conference on artificial intelligence}.

\bibitem[{Kudo(2018)}]{kudo-2018-subword}
Taku Kudo. 2018.
\newblock \href {https://doi.org/10.18653/v1/P18-1007} {Subword regularization:
  Improving neural network translation models with multiple subword
  candidates}.
\newblock In \emph{Proceedings of the 56th Annual Meeting of the Association
  for Computational Linguistics (Volume 1: Long Papers)}, pages 66--75,
  Melbourne, Australia. Association for Computational Linguistics.

\bibitem[{Ma et~al.(2020)Ma, Cui, Si, Liu, Wang, and
  Hu}]{ma-etal-2020-charbert}
Wentao Ma, Yiming Cui, Chenglei Si, Ting Liu, Shijin Wang, and Guoping Hu.
  2020.
\newblock \href {https://doi.org/10.18653/v1/2020.coling-main.4} {{C}har{BERT}:
  Character-aware pre-trained language model}.
\newblock In \emph{Proceedings of the 28th International Conference on
  Computational Linguistics}, pages 39--50, Barcelona, Spain (Online).
  International Committee on Computational Linguistics.

\bibitem[{Mielke et~al.(2021)Mielke, Alyafeai, Salesky, Raffel, Dey, Gall{\'e},
  Raja, Si, Lee, Sagot et~al.}]{mielke2021between}
Sabrina~J Mielke, Zaid Alyafeai, Elizabeth Salesky, Colin Raffel, Manan Dey,
  Matthias Gall{\'e}, Arun Raja, Chenglei Si, Wilson~Y Lee, Beno{\^\i}t Sagot,
  et~al. 2021.
\newblock Between words and characters: A brief history of open-vocabulary
  modeling and tokenization in nlp.
\newblock \emph{arXiv preprint arXiv:2112.10508}.

\bibitem[{Mofijul~Islam et~al.(2022)Mofijul~Islam, Aguilar, Ponnusamy,
  Mathialagan, Ma, and Guo}]{mofijul2022vocabulary}
Md~Mofijul~Islam, Gustavo Aguilar, Pragaash Ponnusamy, Clint~Solomon
  Mathialagan, Chengyuan Ma, and Chenlei Guo. 2022.
\newblock A vocabulary-free multilingual neural tokenizer for end-to-end task
  learning.
\newblock \emph{arXiv e-prints}, pages arXiv--2204.

\bibitem[{Peters et~al.(2018)Peters, Neumann, Iyyer, Gardner, Clark, Lee, and
  Zettlemoyer}]{peters-etal-2018-deep}
Matthew~E. Peters, Mark Neumann, Mohit Iyyer, Matt Gardner, Christopher Clark,
  Kenton Lee, and Luke Zettlemoyer. 2018.
\newblock \href {https://doi.org/10.18653/v1/N18-1202} {Deep contextualized
  word representations}.
\newblock In \emph{Proceedings of the 2018 Conference of the North {A}merican
  Chapter of the Association for Computational Linguistics: Human Language
  Technologies, Volume 1 (Long Papers)}, pages 2227--2237, New Orleans,
  Louisiana. Association for Computational Linguistics.

\bibitem[{Raffel et~al.(2020)Raffel, Shazeer, Roberts, Lee, Narang, Matena,
  Zhou, Li, and Liu}]{raffel2020t5}
Colin Raffel, Noam Shazeer, Adam Roberts, Katherine Lee, Sharan Narang, Michael
  Matena, Yanqi Zhou, Wei Li, and Peter~J. Liu. 2020.
\newblock \href {http://jmlr.org/papers/v21/20-074.html} {Exploring the limits
  of transfer learning with a unified text-to-text transformer}.
\newblock \emph{Journal of Machine Learning Research}, 21(140):1--67.

\bibitem[{Romanov and Shivade(2018)}]{romanov-shivade-2018-lessons}
Alexey Romanov and Chaitanya Shivade. 2018.
\newblock \href {https://doi.org/10.18653/v1/D18-1187} {Lessons from natural
  language inference in the clinical domain}.
\newblock In \emph{Proceedings of the 2018 Conference on Empirical Methods in
  Natural Language Processing}, pages 1586--1596, Brussels, Belgium.
  Association for Computational Linguistics.

\bibitem[{Rust et~al.(2021)Rust, Pfeiffer, Vuli{\'c}, Ruder, and
  Gurevych}]{rust-etal-2021-good}
Phillip Rust, Jonas Pfeiffer, Ivan Vuli{\'c}, Sebastian Ruder, and Iryna
  Gurevych. 2021.
\newblock \href {https://doi.org/10.18653/v1/2021.acl-long.243} {How good is
  your tokenizer? on the monolingual performance of multilingual language
  models}.
\newblock In \emph{Proceedings of the 59th Annual Meeting of the Association
  for Computational Linguistics and the 11th International Joint Conference on
  Natural Language Processing (Volume 1: Long Papers)}, pages 3118--3135,
  Online. Association for Computational Linguistics.

\bibitem[{Sennrich et~al.(2016)Sennrich, Haddow, and
  Birch}]{sennrich-etal-2016-neural}
Rico Sennrich, Barry Haddow, and Alexandra Birch. 2016.
\newblock \href {https://doi.org/10.18653/v1/P16-1162} {Neural machine
  translation of rare words with subword units}.
\newblock In \emph{Proceedings of the 54th Annual Meeting of the Association
  for Computational Linguistics (Volume 1: Long Papers)}, pages 1715--1725,
  Berlin, Germany. Association for Computational Linguistics.

\bibitem[{Sutskever et~al.(2011)Sutskever, Martens, and
  Hinton}]{sutskever2011generating}
Ilya Sutskever, James Martens, and Geoffrey~E Hinton. 2011.
\newblock Generating text with recurrent neural networks.
\newblock In \emph{ICML}.

\bibitem[{Tay et~al.(2021)Tay, Tran, Ruder, Gupta, Chung, Bahri, Qin,
  Baumgartner, Yu, and Metzler}]{tay2021charformer}
Yi~Tay, Vinh~Q Tran, Sebastian Ruder, Jai Gupta, Hyung~Won Chung, Dara Bahri,
  Zhen Qin, Simon Baumgartner, Cong Yu, and Donald Metzler. 2021.
\newblock Charformer: Fast character transformers via gradient-based subword
  tokenization.
\newblock In \emph{International Conference on Learning Representations}.

\bibitem[{Vaswani et~al.(2017)Vaswani, Shazeer, Parmar, Uszkoreit, Jones,
  Gomez, Kaiser, and Polosukhin}]{vaswani2017attention}
Ashish Vaswani, Noam Shazeer, Niki Parmar, Jakob Uszkoreit, Llion Jones,
  Aidan~N Gomez, \L~ukasz Kaiser, and Illia Polosukhin. 2017.
\newblock \href
  {https://proceedings.neurips.cc/paper/2017/file/3f5ee243547dee91fbd053c1c4a845aa-Paper.pdf}
  {Attention is all you need}.
\newblock In \emph{Advances in Neural Information Processing Systems},
  volume~30. Curran Associates, Inc.

\bibitem[{Wang et~al.(2018)Wang, Singh, Michael, Hill, Levy, and
  Bowman}]{wang-etal-2018-glue}
Alex Wang, Amanpreet Singh, Julian Michael, Felix Hill, Omer Levy, and Samuel
  Bowman. 2018.
\newblock \href {https://doi.org/10.18653/v1/W18-5446} {{GLUE}: A multi-task
  benchmark and analysis platform for natural language understanding}.
\newblock In \emph{Proceedings of the 2018 {EMNLP} Workshop {B}lackbox{NLP}:
  Analyzing and Interpreting Neural Networks for {NLP}}, pages 353--355,
  Brussels, Belgium. Association for Computational Linguistics.

\bibitem[{Wu et~al.(2016)Wu, Schuster, Chen, Le, Norouzi, Macherey, Krikun,
  Cao, Gao, Macherey et~al.}]{wu2016google}
Yonghui Wu, Mike Schuster, Zhifeng Chen, Quoc~V Le, Mohammad Norouzi, Wolfgang
  Macherey, Maxim Krikun, Yuan Cao, Qin Gao, Klaus Macherey, et~al. 2016.
\newblock Google's neural machine translation system: Bridging the gap between
  human and machine translation.
\newblock \emph{arXiv preprint arXiv:1609.08144}.

\bibitem[{Wulczyn et~al.(2017)Wulczyn, Thain, and Dixon}]{wulczyn2017ex}
Ellery Wulczyn, Nithum Thain, and Lucas Dixon. 2017.
\newblock Ex machina: Personal attacks seen at scale.
\newblock In \emph{Proceedings of the 26th international conference on world
  wide web}, pages 1391--1399.

\bibitem[{Xue et~al.(2022)Xue, Barua, Constant, Al-Rfou, Narang, Kale, Roberts,
  and Raffel}]{xue2022byt5}
Linting Xue, Aditya Barua, Noah Constant, Rami Al-Rfou, Sharan Narang, Mihir
  Kale, Adam Roberts, and Colin Raffel. 2022.
\newblock Byt5: Towards a token-free future with pre-trained byte-to-byte
  models.
\newblock \emph{Transactions of the Association for Computational Linguistics},
  10:291--306.

\bibitem[{Zhang et~al.(2017)Zhang, Hong, and Balakrishnan}]{poibin_fft}
Man Zhang, Yili Hong, and Narayanaswamy Balakrishnan. 2017.
\newblock \href {https://doi.org/10.48550/ARXIV.1702.01326} {An algorithm for
  computing the distribution function of the generalized poisson-binomial
  distribution}.

\end{thebibliography}
\bibliographystyle{acl_natbib}

\appendix

\section{Improving Pooling Speed}
\label{sec:appendix_cache}
Applying the 1-D convolution requires computing and storing $\mathcal{O}(L_B\times L\times H)$ parameters since we apply the 1D-convolution on every row of the weighted embedding map $P({e^b})^T$. Therefore, this operation may be particularly costly, especially if the frontier predictor outputs a high number of blocks. However, we can use the fact that the weighted embedding map has a special form to reduce the memory load when computing the convolution. Let $K$ be the convolution kernel size, $(C_j)_{j\in [1,K]}\in \mathbb{R}^{K\times H}$ the convolution filters and ``$\cdot$'' denote the element-wise product. Then, omitting padding and biases :
\begin{align*}
    e^B_k &= \max \limits_{i\in [1,L]} \sum \limits_{j=1}^{K} C_j \cdot \left( P_{k,i+j} \cdot e^b_{i+j} \right) \\
          &= \max \limits_{i\in [1,L]} \sum \limits_{j=1}^{K} P_{k,i+j} \cdot \left( C_j \cdot e^b_{i+j} \right)
\end{align*}

Notice how the product between the convolution filters and the byte embeddings $C_j \cdot e^b_{i+j}\in\mathbb{R}^H$ does not depend on the block anymore. We cache this computation, storing $\mathcal{O}(K\times L\times H)$ parameters and only later apply the convolution per block by summing these products with the block-byte membership map $P$. Caching greatly lowers the speed and memory requirements of \ourmodule{}, allowing to save $L_B-1$ element-wise products.\footnote{This caching would be exactly similar if the convolution was not depthwise.} $K$ is usually small, so the products can be stored easily.

\section{Hyperparameters}
\label{sec:appendix_hp}
\begin{table}[H]
\centering\small
\begin{tabular}{lcc}
\toprule
Hyperparameter                                   & $\text{\ourmodule{}}_{Small}$  & $\text{\ourmodule{}}_{Base}$ \\ \midrule
Input Embeddings size          & 64 & 128         \\
Num. layers  & 1 & 2        \\
Num. heads  & 8 & 8        \\
Attention window & 16 & 16        \\
Convolution kernel size & 3 & 3        \\ \bottomrule
\end{tabular}
\caption{Hyperparameters for $\text{\ourmodule{}}$}
\label{tab:hp_fp}
\end{table}

\begin{table}[H]
\centering\small
\begin{tabular}{lccc}
\toprule
Hyperparameter                                   & $\text{ByT5}_{Small}$  & $\text{T5}_{Small}$ & $\text{T5}_{Base}$ \\ \midrule
Hidden size          & 1472 & 512 & 768         \\
Num. layers (encoder)  & 12 & 6 & 12       \\
Num. layers (decoder)  & 4 & 6 & 12     \\
Num. heads  & 6 & 8 & 12      \\
Feed-forward dim. & 3584 & 2048 & 3072        \\
Dropout rate & 0.1 & 0.1 & 0.1      \\ \bottomrule
\end{tabular}
\caption{Hyperparameters for encoder-decoders}
\label{tab:hp_t5}
\end{table}

\clearpage
\onecolumn

\section{Additional results}
We include here the scores obtained by \ourmodel{} on the GLUE test sets for reproducibility and future comparisons. The development sets are used in the main body to allow a fair comparison, as the test scores are not reported in Charformer~\cite{tay2021charformer} and CharBERT~\cite{ma-etal-2020-charbert}.

\begin{table*}[!htp]
\centering\small
\begin{tabular}{cccccccccc}
\toprule
Model                                   & $|\theta|$ & MNLI      & QNLI & MRPC      & SST-2 & QQP       & STSB & COLA & AVG  \\ \midrule
$\text{\ourmodel{}}_{Base}$ (ours)     & 200M        & 78.1/78.2 & 88.6 & 83.6/88.6 & 91.0  & 70.7/88.6 & 74.1 & 45.0 & 78.7\\ [3pt] \bottomrule
\end{tabular}
\caption{Results on test sets for the GLUE benchmark.}
\label{tab:glue_test}
\end{table*}

\section{\ourmodule{} Module}
\label{sec:appendix_detailled}

\begin{figure*}[!htp]
\centering\label{fig:detailled_diagram}
\includegraphics[width=\linewidth]{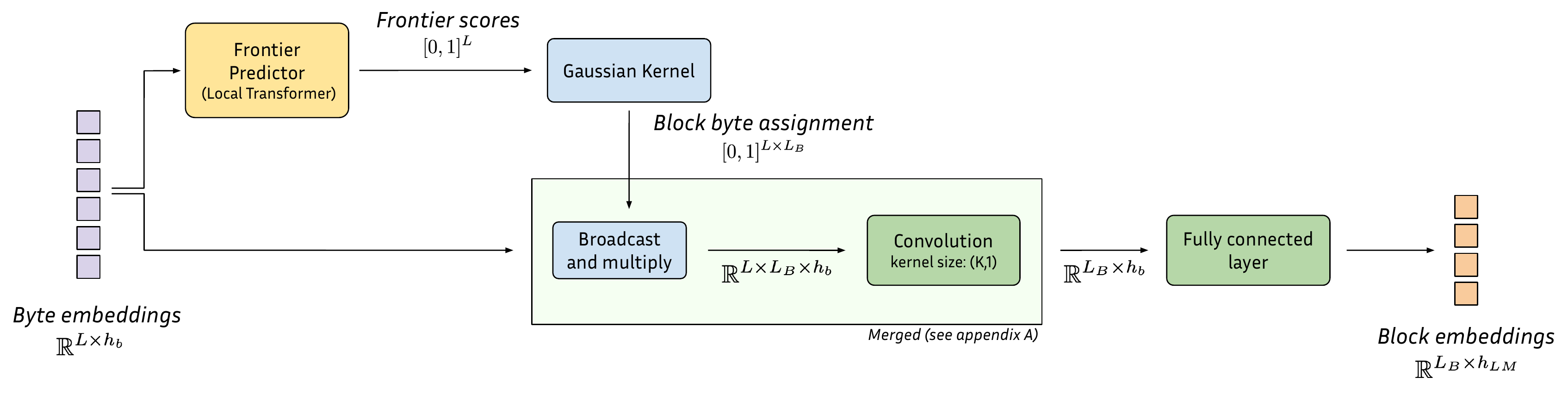}
\caption{A detailed view of the \ourmodule{} module described in section \ref{sec:differentiable_tokenization}. We denote by $h_b$ the dimension of the byte embeddings, by $h_{LM}$ the dimension of the block embeddings that will be fed to the encoder-decoder model, $L$ the length of the input sequence and $L_B$ the length of the block sequence. We omit batch sizes for simplicity.}
\end{figure*}

\end{document}